RESEARCH ARTICLE                                                                   OPEN ACCESS

# Real-Time System of Hand Detection And Gesture Recognition In Cyber Presence Interactive System For E-Learning

Bousaaid Mourad *, Ayaou Tarik *, Afdel Karim*, Estraillier Pascal**
* Department of Computer Science Faculty of Sciences, Ibn Zohr University Agadir, Morocco
** L3i Laboratory University of La Rochelle La Rochelle, France

**ABSTRACT**
The development of technologies of multimedia, linked to that of Internet and democratization of high outflow, has made henceforth E-learning possible for learners being in virtual classes and geographically distributed. The quality and quantity of asynchronous and synchronous communications are the key elements for E-learning success. It is important to have a propitious supervision to reduce the feeling of isolation in E-learning. This feeling of isolation is among the main causes of loss and high rates of stalling in E-learning. The researches to be conducted in this domain aim to bring solutions of convergence coming from real time image for the capture and recognition of hand gestures. These gestures will be analyzed by the system and transformed as indicator of participation. This latter is displayed in the table of performance of the tutor as a curve according to the time. In case of isolation of learner, the indicator of participation will become red and the tutor will be informed of learners with difficulties to participate during learning session.

## I. INTRODUCTION

Gestures, and for long time, have been considered an interactive technique, which may potentially offer methods more natural, creative and intuitive to communicate via Human Computer Interfaces. Recently, many searches have been carried out to develop Human Computer Interfaces using hand gesture [5] [6] [9]. These searches have come up with important and impressive results concerning recognition of hand gesture based on detecting hand movement and its position [7]. In E-learning context, the quality and quantity of asynchronous and synchronous communications are the key elements for Distance learning success. It is essential to offer a propitious supervision to reduce isolation feeling among E-learners, as such feeling is considered to be one of main direct causes of loss and high rates of disengagement in distance learning. To avoid this isolation feeling among E-learners, we are aiming, by this work, to provide trainers and learners with an environment permitting them to behave as being in real class. One of the privileged means to be adopted is the use of USB cameras to analyze, in real time, movements linked to trainer and learners behaviors. Besides, this work aims the recognition of learners' "feedback" quality reactions by the trainer like those he is accustomed to in real class, especially learners' attention and their level of participation. In order to conceive an environment of production and personalization of pedagogical interactive applications, the searches, we are carrying out, aim to provide convergent solutions coming from real-time image, non-intrusively captured, implicit and explicit behaviors of trainer and learners based on hand gesture. The latter will be analyzed by the system and transformed as indicator of participation. This indicator is displayed in the tutor's dashboard as a curve according to the time. The indicator is defined as the frequency of raising hand of learner to participate during a learning session. In case of isolation of learner, the indicator of participation will become red and the tutor will be informed of learners with difficulties to participate during learning session. In this paper, we are going to define in the first part the System architecture for the detection and recognition of hand gesture. The second part will deal with hand gesture detection with codebook, Haar Cascade method and Camshift method. The third party will tackle recognition of hand gestures with contour point distribution histogram (CPDH), and we will end by conclusion.

## II. SYSTEM ARCHITECTURE FOR THE DETECTION AND RECOGNITION OF HAND GESTURE

Hand gesture is one of the most important tools for communication in human daily life. For instance, while learning, such gesture can be used to communicate or stop the tutor if learner does not understand some concepts, or he wants to answer questions intermittently posed by the tutor. These gestures can be quantified as an indicator of learner attention and his level of participation during a learning session.

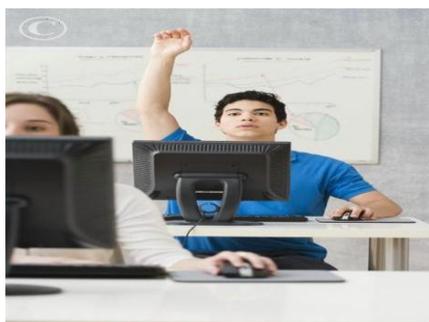

Fig 1: Use of hand gesture as an indicator of participation

This system is composed of following stages:
- Stage of processing and detection of hand gesture;
- Stage of hand gesture recognition;

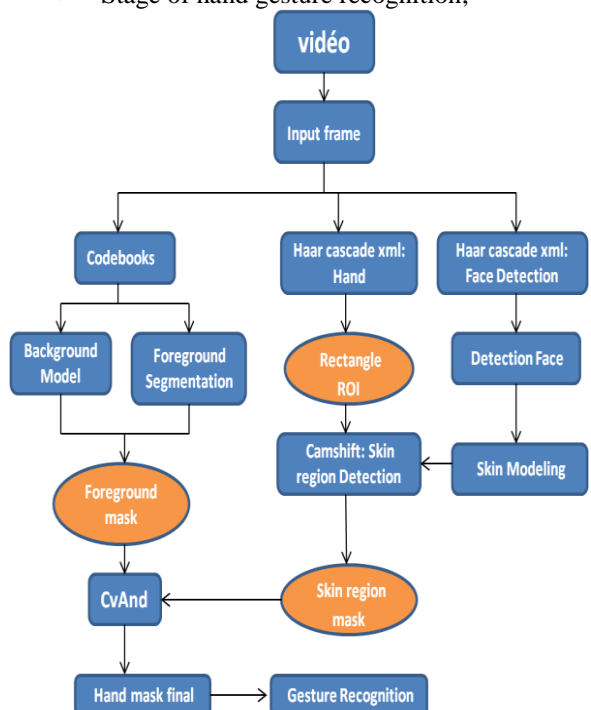

Fig 2: Implementation of the detection system, the recognition of hand gesture

### III. HAND GESTURE DETECTION

To be able to detect hand gesture, there are many processes tot be carried out so as to eliminate information which may damage this stage of detection like light variation and problems of static and dynamic video sequences background:

#### 1- *Elimination of noise video sequences background:*

In this stage, we eliminate useless information of compound video sequences background and problems of light changes. For this purpose, we use Codebook method to construct static model background [2]. The base idea of this method consists of monitoring pixels evolution in color space YCbCr and representing their variations in Codebook. In this stage of learning, created [3] 90 image sequences are used to construct static model background [1]. We extract image of hand gesture by movement of previous constructed static model background. This image, then, is adopted to binary system and processed by morphological operators to obtain first binary mask of moving hand gesture.

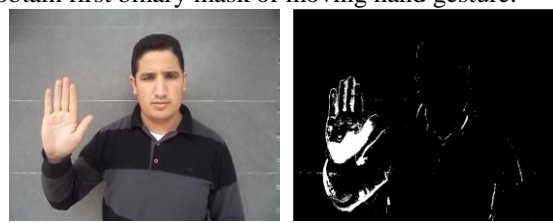

(Input frame)         (Foreground mask)

Fig 3: examples of first binary mask of moving hand gesture with codebook

#### 2- *Detection of the hand based on Haar Cascade method:*

Haar Cascade is a method of image object detection, proposed by searchers Paul Viola and Michael Jones [10]. It is a part of all first methods capable of detecting effectively and in real time image objects. Originally invented to detect faces [10], it may also be used to detect other types of objects like cars or planes. In our work, we have created our own haar Cascade to detect the hand. Detection of an object using haar Cascade classifier can be divided into:

- Creation of file description of positive samples constituted here by hand gestures
- Creation of file description of negative samples by objects other than hand gesture. Here to avoid detection of face skin, we may put images of face as negative samples
- Packaging of positive samples in a file
- Stage of classifier learning
- Conversion of learning cascade in xml file
- Use of xml file to detect hand gestures.

The use of this method leads to relevant results in hand detection, but it has a slow disadvantage because of the learning phase, this time is disadvantageous when making the detection in real time. For this, we will use this method to locate the position of the hand (ROI). This latter is transmitted to Camshift allowing the detection and tracking of the hand. This method is to be used periodically after every T = 20 frames to initialize Camshift function not to lose the tracking of the hand.

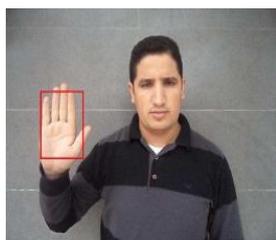

Fig 4: hand detection using our Haar cascade xml: Hand

Once the hand gesture is detected, we can provide coordinates of rectangle surrounding hand gesture or what we call ROI, which will be the next initial stage that will locate hand position for CamShift function.

### 3- Detection of skin by CamShift technique:

To detect the skin, we use Camshift method. This technique uses the skin color to construct the second mask of hand gesture movement. The problem of this technique is its sensitivity to light variation. To solve this problem and facilitate skin detection, a part of the face is chosen using Haar Cascade to detect the face of Jones and Viola [10]. Our purpose is selecting the skin pixels, then calculating statistics on adaptable distribution of skin color and defining also the model representing the skin color [8][11]. After thresholding and adequate morphological processing, we can obtain from this model a second mask of hand gesture movement. We will combine the two masks obtained on the same frames using the "and" operator to remove the pixels corresponding to the face and keep only the pixels belonging to the hand.

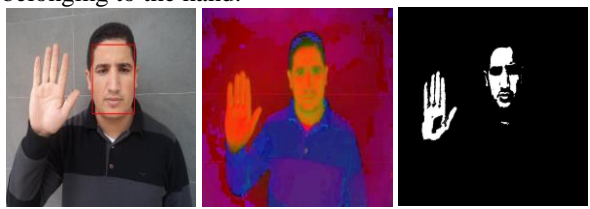

(a)   (b)   (c)

Fig 5: a: face detection with harr cascade face, (b): HSV model, (c): Skin region Detection mask after thresholding and adequate morphological processing

## IV. RECOGNITION OF HAND GESTURES WITH SHAPE FEATURE CONTOUR POINT DISTRIBUTION HISTOGRAM (CPDH)

Once we got the mask corresponding to the hand gesture, we will move to the stage of gesture recognition. For this, we use shape feature CPDH technique [4]. This technique is based on the distribution of points on object contour under polar-coordinates. In this algorithm, the object boundary is detected using standard Canny operator to describe its shape. The result points on the contour can be represented as,

$P = \{(x_1, y_1), (x_2, y_2), \dots, (x_n, y_n)\}, (x_i, y_i) \in R^2$,

where n represents the total number of points on contour. After the extraction of points on object boundary, the centroid of the hand boundary is computed. The maximum value of distance between the centroid and the boundary points on the contour is chosen as the radius of the minimum circumscribed circle.

The summarized algorithm for the construction of CPDH [4] is given as:

- Step 1: Input a binary shape image.
- Step 2: Extract object contour points with Canny operator.
- Step 3: Sample out N points on the contour with x and y coordinates:
- $P = \{(x_1, y_1), (x_2, y_2), \dots, (x_n, y_n)\}, (x_i, y_i) \in R^2$.
- Step 4: Compute the centroid of the shape $(x_c, y_c)$.
- Step 5: Set the centroid as the origin and translate P into polar coordinates,
- $P = \{(\rho_1, \theta_1), (\rho_2, \theta_2), \dots, (\rho_n, \theta_n)\}, (\rho_i, \theta_i) \in R^2$. where $\rho_i = \sqrt{(x_i - x_c)^2 + (y_i - y_c)^2}$ is the distance between the points $(x_i, y_i)$ and $(x_c, y_c)$
- $\theta_i = \arctan((y_i - y_c)/(x_i - x_c))$ is the angle between $\rho_i$ and x-axis.
- Step 6: Get the minimum circumscribed circle C with the centre $(x_c, y_c)$ and radius $\rho_{max}$, where $\rho_{max} = \max\{\rho_i\}$, i=1,2,…..,n.
- Step 7: Partition the area of C into $u \times v$ bins with u bins for $\rho_{max}$ and v bins for $\theta$.
- Step 8: Construct the CPDH of the shape image by counting the number of points which are located in every bin.
- Step 9: Output CPDH.

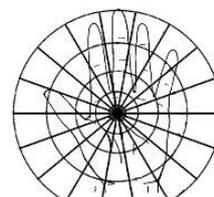

Fig 6: CPDH shape feature

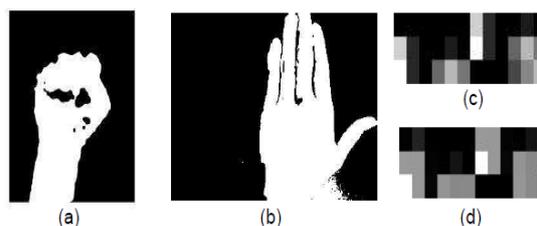

Figure 7: (a) and (c): binary mask hand, (b) and (d): a CPDH's constructor correspondent.

## V. EXPERIMENTAL RESULTS

The results obtained in our work presented below in figure 8. We show in image (c) that the adaptive Camshift method leads to good results in detecting the skin, after the dynamic and adaptive modeling of skin color. However, this technique does not only detect the hand, but the face as well. The image (d) shows the result of using codebook to extract background. To get only the mask corresponding to the gesture of the hand, we combine the two masks obtained from both techniques. The image (e) shows the result of our basic idea of combining the two masks to get the final one. The latter mask is then post-processed for better results. Furthermore, In order to remove noisy and blurry image, morphological operations such as opening and closing are applied followed by Gaussian blurring and thresholding.

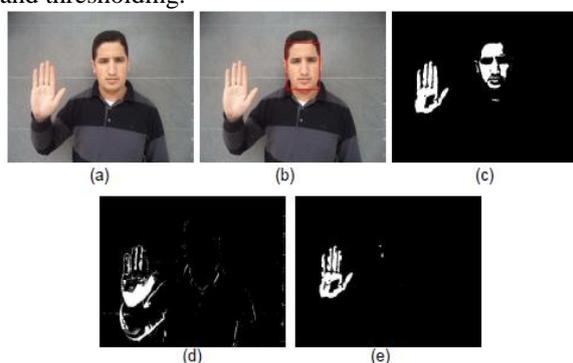

Fig 8: (a) original Image, (b): Face detection using our haar cascade, (c): Skin region Detection mask, (d): Foreground mask with codebook, (e) Hand mask final with CvAnd and adequate morphological processing.

Concerning gesture recognition, we use shape descriptor Contour Point Distribution Histogram CPDH. We limit ourselves to two types of hand gestures: Palm and Fist. We use a database of 200 images of hand gestures and calculate the CPDH for each of these images to get finally a CPDH database of hand gestures. The latter will be relied on to decipher the type of hand gesture. That is, if the gesture is a palm or fist. To achieve this, we calculate the CPDH hand gesture obtained in the detection step and compare it to the CPDH database of hand gesture, then get the closest one by using Euclidean distance and determine, thereafter, the type of hand gesture.

The rate of accuracy and performance of the proposed approach is measured by applying real database for both hand gestures, namely palm and fist. The evaluation of the performance of our system is based on the values of recall and precision, which are defined as follows:

$$recall = \frac{true\ positive\ detected\ (TP)}{total\ true\ positves} \times 100\%$$

$$precision = \frac{true\ positive\ detected\ (TP)}{all\ detections} \times 100\%$$

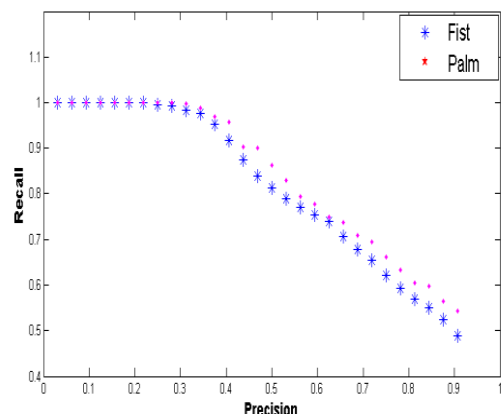

Fig 9: the ROC curve for the recognition of two gestures Palm and Fist.

## VI. CONCLUSION

The objective of this research is to provide the trainer and every learner with an environment that allows them to behave as if being in real class. One of the preferred ways is the use of USB cameras for hand gesture analysis in real-time. For this end, we have implemented a system of detection and hand gesture recognition in order to use it as an indicator of participation. We started by applying the codebook method to create the first binary mask of moving hand. In order to get the second mask, we incorporated color-based detection step. We used a robust approach for face detection by method of Jones and viola, where human skin areas are estimated by using skin color model. The skin color distribution in histogram map gives an idea to choose appropriate thresholds values for detecting skin pixels. These values are transmitted by CamShift to get the skin region mask. Finally, we combine the foreground and the skin masks to get the final mask for hand region. The obtained mask is then post-processed for better results. In order to remove noisy and blurry image, morphological operations, such as opening and closing, are applied. Finally, in the hand gesture recognition phase, we have used the CPDH technique. We started with the recognition for both hand gestures, namely palm and fist, to be used as an indicator of participation.

## REFERENCES
**Journal Papers:**